\documentclass{article} 
\usepackage{iclr2027_conference,times}

\usepackage{amsmath,amsfonts,bm}

\def\eqref#1{equation~\ref{#1}}

\def\1{\bm{1}}

\DeclareMathAlphabet{\mathsfit}{\encodingdefault}{\sfdefault}{m}{sl}
\SetMathAlphabet{\mathsfit}{bold}{\encodingdefault}{\sfdefault}{bx}{n}

\usepackage{hyperref}
\hypersetup{hidelinks}
\usepackage{url,graphicx,booktabs,multirow,makecell,colortbl,xcolor}
\usepackage[table]{xcolor}
\usepackage{pifont}
\usepackage{fontawesome5}

\definecolor{audiobg}{HTML}{EFEFEF}

\title{Full-duplex Spoken Language Modeling with Discrete Diffusion}

\author{Chen Chen \ \  Pin-Jui Ku \ \ Ankita Pasad \ \  Ante Jukić \ \ Zhehuai Chen \ \ Boris Ginsburg  \\
NVIDIA\\ \\
}

\begin{document}

\iclrfinalcopy
\maketitle

\begin{abstract}
We introduce \textbf{\textit{Diffuplex}}, a full-duplex spoken language model based on discrete diffusion that achieves an unprecedented balance between linguistic intelligence and response efficiency. With thinking mode, Diffuplex delivers $13.0$ tokens per forward pass, reaching an average of $1872$ tokens per second and substantially surpassing the frontier autoregressive (AR) models in decoding speed. Without thinking, Diffuplex matches the first-audio latency of existing full-duplex models while achieving markedly stronger linguistic intelligence. This capability is enabled by separating \emph{when to say} from \emph{what to say} at the training level, rather than compressing both into a single autoregressive timeline. During inference, Diffuplex continuously tracks the user speech through an AR interaction policy, while independently generating and refining response content through parallel diffusion. Consequently, semantic computation is no longer bounded by the perception audio frame rate, allowing intelligence and responsiveness to improve simultaneously. Taken together, these results establish a new paradigm for spoken agents that can sustain real-time dialogue while devoting parallel computation to intelligent response generation. Training code and checkpoints will be avaliable soon\footnote{https://cchen1436.github.io/diffuplex-website}.
\end{abstract}

\begin{figure*}[h]
    \centering
    \includegraphics[
        width=1.0\textwidth
    ]{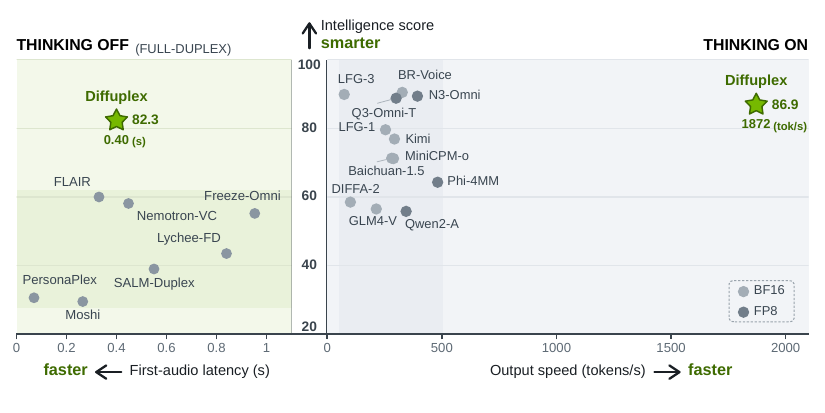}
   \caption{\textbf{Spoken intelligence and response efficiency.} The shared vertical axis is the VoiceBench overall score across 9 subsets, higher score means smarter model. \textit{Left:} Diffuplex without thinking versus full-duplex spoken dialogue models, plotted against first-audio latency (from the end of user speech to the start of model speech). \textit{Right:} Diffuplex with thinking versus frontier turn-based AR models, plotted against output speed. All output speeds were measured at batch~1 on the same NVIDIA GB300 GPU. More decoding speed results on different GPUs are provided at Appendix~\ref{app:fig1-details}.}
    \label{fig:two_axes}
\end{figure*}

\section{Introduction}

An ideal spoken language model should exhibit both fluid conversational behavior (\emph{when to say}) and intelligent response generation (\emph{what to say}). From a cognitive perspective, \emph{when to say} requires tracking real-time speaking stream of interlocutor, whereas \emph{what to say} involves progressively formulating internal thoughts. Neural recordings during conversation identify distinct activity associated with speech perception, planning, and production, even when these processes overlap in time~\citep{castellucci2022speechplanning}. This suggests a useful design principle: conversational timing and response formulation should be coordinated without being forced into a single sequential process.\par

Turning this separation into a spoken language model, however, remains a formidable challenge. While recent full-duplex models that unify when to say and what to say deliver fluid interactions, they exhibit severe linguistic and cognitive degradation compared with frontier autoregressive (AR) models---which handle only \emph{what to say} and rely on an external ‌voice activity detection to govern response timing. We argue that this degradation arises from conflating these two learnable dimensions into a single one-dimensional AR timeline: (1) Conflicting Objectives: Optimizing both tasks via a single Cross-Entropy loss results in gradient conflict, where the dense supervision of low-level temporal alignment heavily dilutes the delicate representations required for high-level semantic generation~\cite{li2026decoupling}. (2) Temporal Frame-Rate Binding: By tying AR generation directly to a rigid physical frame rate, e.g., 80 ms in Moshi~\citep{defossez2024moshi}, the model is compelled to process semantics in lockstep with real-time spoken input, stripping its capacity to allocate elastic test-time compute (i.e., Chain-of-Thought) for deep inference. \par

\begin{figure*}[t]
    \centering
    \includegraphics[
        width=0.99\textwidth
    ]{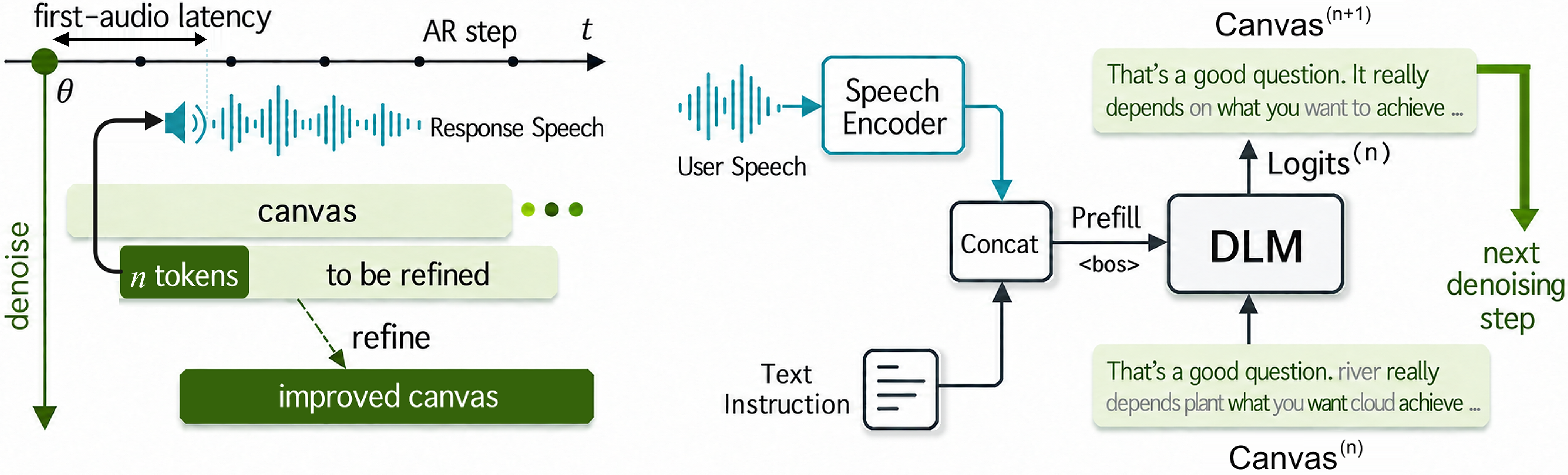}
    \caption{
    Overview of Diffuplex. \textbf{Left:} The horizontal axis represents real-time AR interaction, while a predicted \texttt{<bos>} triggers diffusion generation along the vertical axis at $\theta$. An early refined prefix of $n$ tokens is streamed to the speech synthesizer as the remaining canvas continues denoising. \textbf{Right:} Speech features and the text instruction are jointly prefilled into the DLM as condition, which iteratively refines each response canvas through bidirectional denoising. 
    }
    \label{fig2}
\end{figure*}

To break this coupling, we exploit a distinctive property of discrete diffusion language models (DLM): response tokens can be generated and refined in parallel~\citep{nie2025llada}. This introduces a second computational axis that is independent of physical time. As illustrated in Figure~\ref{fig2}, an AR policy advances along the horizontal timeline $t$, continuously processing incoming speech and determining when a response should begin. Once triggered at time $\theta$, a diffusion process expands orthogonally from the timeline and iteratively refines an entire response canvas (i.e., a generation block). To support streaming interaction, we further design an anchored first-$n$ decoding strategy: the first $n$ response tokens are finalized early and immediately synthesized, while audio playback provides a time buffer for the remaining token refinement.  \par

We instantiate this two-axis formulation as \textbf{\textit{Diffuplex}}, a spoken language model that separates semantic generation from conversational control within a unified backbone. We efficiently connect a pretrained speech encoder~\citep{nvidia2025parakeet} to a text DLM~\citep{diffusiongemma2026}, allowing Diffuplex to take spoken input as diffusion condition, while inheriting the linguistic capabilities and knowledge encoded in the text model. Since the same backbone can also operate in AR mode, we introduce only a lightweight state embedding and prediction head to learn semantically informed speaking decisions from conversational data. At inference time, the two modes assume complementary roles: the AR mode continuously processes the incoming speech stream and determines \emph{when to say}, while the diffusion mode is triggered at appropriate moments to generate and iteratively refine \emph{what to say} in parallel. \par

Experiments on public benchmarks evaluate Diffuplex along two complementary dimensions: response quality and conversational behavior. With thinking enabled, Diffuplex achieves an overall score of 86.9 across the 9 VoiceBench subsets~\citep{chen2024voicebench}, approaching the strongest leaderboard systems while finalizing an average of 13.0 tokens per forward pass, compared with one token for the evaluated AR baselines. Meanwhile, Diffuplex exhibits responsive full-duplex behavior, with turn-taking and barge-in latency comparable to strong full-duplex models, while exceeding existing open-source systems by more than 20 points in spoken intelligence. \par

Our contributions are threefold. First, we show that diffusion-based spoken language modeling delivers both strong linguistic capabilities and exceptional generation speed, averaging 1,872 tokens per second on a single GB300 GPU—far outpacing its AR counterparts. Second, we demonstrate that conversational control and response generation can be learned separately within a shared backbone: an AR policy tracks streaming speech, while diffusion generates response content. Third, we show how to translate fast parallel text generation into full-duplex spoken interaction, enabling rapid responses to conversational events such as turn-taking and barge-in. We believe these contributions lay the groundwork for building both fast and intelligent spoken agents, paving the way for seamless and highly capable voice interactions.

\section{Related Work}
\subsection{Spoken Language Models}
Turn-based spoken language models extend text LLMs with speech encoders or discrete acoustic tokens, enabling unified speech understanding and generation \citep{wang2026towards}. AudioPaLM transfers linguistic knowledge from a text LLM to speech tasks, while SpeechGPT develops cross-modal instruction-following through modality adaptation and speech instruction tuning \citep{rubenstein2023audiopalm,zhang2023speechgpt}. More recent systems, such as Qwen2.5-Omni, jointly generate text and speech through specialized reasoning and synthesis modules \citep{xu2025qwen25omni}. Despite their strong linguistic capabilities, these models largely retain a turn-oriented interaction protocol and generate response content autoregressively.

Full-duplex models instead process user and agent speech as concurrent streams, allowing interruptions, overlaps, and backchannels without explicit turn segmentation \citep{wang2026end}. Moshi models the two speakers with parallel audio-token streams, while OmniFlatten serializes multimodal dialogue events into a unified sequence \citep{defossez2024moshi,zhang2024omniflatten}. Freeze-Omni further connects streaming speech interfaces to a frozen text LLM to preserve its linguistic capabilities \citep{wang2024freezeomni}. PersonaPlex adds role and voice control to a full-duplex speech model, Lychee-FD addresses acoustic--semantic interference, and Listen-Write-Speak provides a visible text channel during real-time conversation~\citep{roy2026personaplex,liu2026lychee,zhang2026lws}. These systems achieve responsive interaction, but typically bind semantic generation and conversational control to the same frame-synchronous AR process. Diffuplex instead separates these functions: an AR policy tracks conversational timing, whereas a DLM generates response content along an independent computational axis.

\subsection{Diffusion Language Models with Spoken Input}

Diffusion language models generate text through iterative refinement rather than strictly left-to-right decoding. Early work explored continuous diffusion for controllable text generation \citep{li2022diffusionlm}, while masked discrete diffusion substantially narrowed the likelihood gap with AR models \citep{sahoo2024mdlm}. LLaDA demonstrated that diffusion language models can scale to general-purpose pretraining and instruction following \citep{nie2025llada}. Most relevant to our work, DiffusionGemma converts a pretrained AR mixture-of-experts model into a discrete DLM that refines 256-token canvases in parallel while retaining AR capabilities \citep{diffusiongemma2026}.

The extension of DLMs to spoken input remains comparatively underexplored. Recent works have applied diffusion decoders to
automatic speech recognition (ASR), either training a speech-conditioned diffusion decoder from scratch or attaching a frozen encoder or DLM for parallel transcript decoding~\citep{baas2022transfusion,kwon2025whisfusion,yen2026mdmasr,khurdula2026audionative}. DIFFA moves beyond transcription by first aligning speech through ASR and subsequently training speech-conditioned instruction following; DIFFA-2 further scales this framework to general audio understanding \citep{zhou2026diffa,zhou2026diffa2}. These studies establish that DLMs can understand spoken input, but do not address full-duplex interaction or explicitly separate conversational timing from response generation. Diffuplex closes this gap by combining a streaming AR interaction policy with parallel diffusion-based response generation. Cross-modal distillation has also been explored to align the capabilities of speech models with text teachers~\citep{cao2026xopd}.
\section{Method}
\label{sec:method}

\textbf{Overview.} Diffuplex follows a three-stage training curriculum. In Stage~1, we perform speech--text modality alignment using ASR data, enabling a pretrained text diffusion language model to condition on continuous speech representations. In Stage~2, we train the model on spoken-instruction--text-response pairs to acquire instruction-following capabilities, thereby learning \emph{what to say}. In Stage~3, we preserve the capabilities acquired in Stage~2 while introducing an AR interaction mode that learns a conversational policy, i.e., \emph{when to say}.

At inference time, the model operates along two decoupled computational axes. A causal AR branch continuously processes incoming speech to model conversational behavior, while a diffusion branch generates responses conditioned on the accumulated context. In Section~\ref{infer_and_denoise}, we introduce denoising strategies designed to enable low-latency and fluid speech-to-speech interaction.
\subsection{Stage 1: Modality Alignment}
We construct the speech frontend using a Parakeet encoder~\citep{nvidia2025parakeet} followed by a randomly initialized projection layer that maps acoustic representations into the embedding space of DiffusionGemma~\citep{diffusiongemma2026}. Given these speech embeddings and textual instruction as conditioning context, the model reconstructs the complete ASR transcript through iterative denoising. We first warm up the projection layer on LibriSpeech~\citep{panayotov2015librispeech}, then jointly optimize the projection layer and speech encoder on an English ASR mixture drawn primarily from Granary~\citep{raokoluguri2025granary}. The diffusion language model remains frozen throughout this stage, requiring the speech frontend to produce representations compatible with its pretrained textual latent space.

\paragraph{Auxiliary loss.} Recent work suggests that an auxiliary CTC objective can improve diffusion-based ASR performance~\citep{khurdula2026audionative}. We did not observe consistent gains in our setting and instead employ an auxiliary AR next-token loss alongside the diffusion objective. The resulting Stage 1 model achieves a word error rate (WER) of $2.3$ on LibriSpeech~\citep{panayotov2015librispeech} test-clean and $4.2$ on test-other. 

\subsection{Stage 2: Spoken Instruction-following}
We sample 600K speech-instruction examples from the Nemotron-Omni SFT corpus~\citep{nvidia2026nemotron3omni} and mix them with 5\% ASR replay, drawn from 37.5K transcription examples. To strengthen the capability without thinking, 70\% of responses directly supervise final answers without explicit \texttt{<think>} content. The speech encoder, projection layer, and DLM are jointly optimized using the same diffusion and auxiliary AR objectives as in Stage~1, while the MoE router remains frozen. \par
Joint adaptation exposes an empirical trade-off. With a small learning rate, long-form generation of DLM remains coherent, but the model only weakly grounds its responses in the spoken input. Increasing the learning rate improves speech adaptation but causes the pretrained language distribution to drift, resulting in repetition and incorrect termination. This tension cannot be addressed by setting different learning rates for three modules, or replaying text-only data.

\paragraph{Text-Anchored distillation.}
We therefore retain a frozen copy of the pretrained DLM as a textual anchor. For each paired training example, the trainable student receives the spoken request, whereas the frozen teacher receives its written counterpart. Then we minimize forward KL divergence between their output distributions for both the diffusion canvas and the causal AR readout. This cross-modal supervision effectively transfers the anchor teacher's textual capabilities to the speech-conditioned student.

\subsection{Stage 3: Full-duplex Interaction Modeling}
Natural conversation does not arrive as a sequence of pre-segmented turns with explicit boundaries. We therefore learn a streaming interaction policy directly from conversational speech data. The user audio is treated as a continuous stream and divided into 80-ms chunks. At each step, the model autoregressively predicts one interaction state conditioned only on the observed speech prefix and previous states.

In Diffuplex, we represent this policy using four states:
\texttt{<listen>}, \texttt{<speak>}, \texttt{<bos>}, \texttt{<eos>}. The first two indicate whether the model should listen or retain the conversational floor, whereas \texttt{<bos>} and \texttt{<eos>} denote taking and yielding the floor, respectively. These transition events are sparse and therefore receive larger loss weights than the persistent states:
\begin{equation}
\mathcal{L}_{\mathrm{policy}}
=
-\sum_t w_{s_t}
\log p(s_t \mid a_{\leq t}, s_{<t}),
\end{equation}
where \(a_{\leq t}\) denotes the speech observed up to time \(t\). As the model cannot anticipate user's behavior, the \texttt{<bos>} and \texttt{<eos>} targets include a short reaction delay that allows sufficient acoustic evidence to accumulate. The interaction state tokens are independent of the DLM vocabulary, with a trainable word embedding matrix. At each AR step, the previous state embedding is combined with the chunk-level speech representation for the next state token prediction. \par

\textbf{Causality of encoder.} Stage 3 reuses speech encoder from Stage 2 but operates under strict causal constraints: a 10-second trailing window without lookahead aligned with streaming inference. To bridge this representation shift, a lightweight LoRA adapter~\citep{hu2021lora} is introduced, but is not activated when computing diffusion context. This design preserves encoding quality while meeting online constraints, without requiring a separate causal encoder or suffering train-test mismatch.

\subsection{Inference and Denoise} \label{infer_and_denoise}
When the AR branch emits \texttt{<bos>}, the speech-encoder features accumulated so far are prefilled, together with the text prompt, through the DLM's causal pass to build its key--value cache, and the response is generated as a sequence of 256-token canvases that attend to this cache. Each denoising pass is one bidirectional forward over the canvas. We largely follow the inference recipe of DiffusionGemma~\citep{diffusiongemma2026}: at every pass its entropy-bounded sampler selects the lowest-entropy positions whose mutual-information bound stays below an entropy bound $\beta{=}0.1$, keeps their sampled tokens for the next pass, and fully re-noises all non-selected positions; the sampling temperature decays linearly across the budget of at most $N{=}48$ passes. Denoising terminates early once the highest-probability predictions are identical across two consecutive passes and the average entropy over the canvas falls below a confidence threshold $\gamma$. The converged canvas is committed and appended to the cache before the next one is opened. Because the original recipe was designed for offline text generation, we depart from it in two respects that are essential for spoken interaction. First, the opening of the response is committed long before the canvas converges, so that speech synthesis can start within a few passes (\emph{anchored prefix}). Second, the denoising schedule is shortened so that fewer passes separate consecutive speech packets (\emph{denoising schedule}).

\paragraph{Anchored prefix.} Under the original decoding scenario, the first word of the answer is final only when the whole canvas has converged, typically after tens of passes. The leading positions, however, settle much earlier. We therefore commit the first $n{=}8$ positions of the first canvas after at most $K$ passes and stream them to the synthesizer while the rest keeps denoising. Position~0 is seeded with the argmax of the causal AR readout at the prefill, which is trained with the auxiliary AR loss and costs nothing extra. During the first $K{-}1$ passes, non-accepted front positions are refilled with the denoiser's argmax rather than noise, and a front position $i<n$ is frozen at its argmax at the first pass $j$ satisfying
\begin{equation}
H\bigl(p_i^{(j)}\bigr)<\tau \quad\text{or}\quad j=K ,
\end{equation}
where $p_i^{(j)}$ is the temperature-scaled distribution at pass $j$ and $\tau{=}0.05$. Frozen positions are never re-sampled, so the first $n$ tokens can be handed to the TTS system after one prefill and at most $K$ passes regardless of response length, and speech synthesis begins while the remainder of the canvas is still denoising; $K$ trades latency against the quality of the opening (Appendix~\ref{app:prefix-sensitivity}).

\paragraph{Denoising schedule.} The pretrained schedule (temperature $0.8\!\to\!0.4$, $\gamma{=}0.005$) spends its last passes confirming an already converged canvas. We anneal from $0.7$ to $0.2$ and relax $\gamma$ to $0.025$, and cut any response whose canvas ends in a trailing run of $\ge 32$ tokens drawn from at most two distinct tokens, a degenerate loop that a sharper schedule occasionally produces. Committed text is streamed to an interleaved text--speech TTS model~\citep{du2025cosyvoice3}, the anchored prefix forming the first packet, so synthesis of the opening overlaps with the denoising of the remainder. \par
The AR interaction policy and diffusion response generator share the same backbone parameters but maintain separate KV caches. The AR cache advances causally with incoming audio to support turn-taking decisions; once the policy emits \texttt{<bos>}, the observed speech context is separately prefilled to initialize the diffusion cache. This keeps the policy’s streaming history distinct from the context used to denoise the response.

\subsection{Speech Synthesizer}
As the first $n$ tokens are finalized early in the denoising process, Diffuplex is agnostic to the choice of speech synthesizer. These tokens can be immediately passed to any high-quality TTS model while the remaining canvas continues to be refined; consequently, the TTS model itself need not provide native streaming inference. In this work, we adopt CosyVoice~3~\citep{du2025cosyvoice3} for its synthesis quality and voice-cloning capability. When voice cloning is unnecessary, it can be replaced with a smaller single-speaker TTS model to further reduce first-audio latency. This modular interface allows synthesis quality, speaker flexibility, and latency to be traded off without retraining Diffuplex. \par
\textbf{User interruption.} When the AR interaction policy emits \texttt{<eos>}, any remaining response audio is immediately truncated, allowing Diffuplex to yield naturally to the user. If the user interrupts within one second, we discard this unspoken canvas rather than appending it to the KV cache for the next interaction. Consequently, only content that has effectively entered the conversation is retained in the context, preventing frequent interruptions or hesitations from corrupting conversation context.

\section{Experimental Setup}

\subsection{Training Details}
Training follows the three stages described in Section~\ref{sec:method}. Appendix~\ref{app:training-details} summarizes the training data, configurations, computing resource, and optimization settings for all stages.

\subsection{Benchmark and Evaluation}
\textbf{VoiceBench}~\citep{chen2024voicebench} evaluates the spoken intelligence of voice assistants across open-ended dialogue, knowledge and reasoning, instruction following, and safety. It comprises nine evaluation subsets with both real and synthetic speech, incorporating variations in speakers, acoustic environments, and linguistic content. The metric contains both LLM judge score and accuracy, and we report them on average for comparison.  \par
\textbf{Full-Duplex-Bench v1}~\citep{lin2025fullduplexbench} evaluates fundamental real-time conversational behaviors under four controlled scenarios: pause handling, backchanneling, smooth turn-taking, and user interruption. We also add the ``turn-taking dynamics" track from \textbf{Full-Duplex-Bench v3}~\citep{lin2026fullduplexbenchv3}, as all recording is collected from human speakers with conversational disfluencies.  \par

\subsection{Baseline}
We compare Diffuplex against three classes of spoken models: (1) \textbf{Turn-based AR models} represent the dominant high-capability paradigm, and we enable thinking where supported. As the intermediate reasoning trace is not presented to the user, enabling thinking in AR models would result in relatively high first-audio latency. (2) \textbf{Turn-based diffusion models} generate response tokens through parallel iterative refinement and can therefore reduce decoding latency relative to autoregressive generation. (3) \textbf{Full-duplex AR models} interleave speech perception and generation along a fixed-rate temporal stream, enabling low first-audio latency and fluid turn-taking. Because AR steps are tied to perceived audio chunks, these systems do not have a separate thinking phase. Original publications and model cards for every compared system are listed in Appendix~\ref{app:model-sources}. \par

\section{Evaluation and Results}
\label{sec:results}

\subsection{Results on Spoken Intelligence}
\label{sec:spoken_intelligence}
Table~\ref{tab:voicebench} reports two evaluation settings for Diffuplex. Stage~2 evaluates the speech-conditioned response generator on complete spoken requests, with and without thinking. Stage~3 starts from the Stage~2 checkpoint, freezes the response generator, and adds the streaming interaction policy; its row evaluates the full-duplex system by committing the first $n{=}8$ tokens within $K{=}4$ denoising passes. Results marked with $^{*}$ were reproduced by us when the corresponding numbers were unavailable in the source paper~\citep{zhou2026diffa2}.

\begin{table*}[t]
\centering
\caption{
VoiceBench results across turn-based AR and full-duplex models. Open-ended QA columns use a 5-point judge scale; the remaining columns and Overall use a 0--100 scale.
}
\label{tab:voicebench}
\setlength{\tabcolsep}{4.5pt}
\resizebox{\textwidth}{!}{
\begin{tabular}{l|ccc|cccc|cc|c}
\toprule
\multirow{2}{*}{Model}
& \multicolumn{3}{c|}{Open-ended QA}
& \multicolumn{4}{c|}{Knowledge and Reasoning}
& \multicolumn{2}{c|}{Ins. and Safety}
& \multirow{2}{*}{\textbf{Overall}} \\
& Alp. & Com. & Wil.
& SDQA & MMSU & OBQA & BBH
& IFE. & Adv. & \\ \midrule

\rowcolor{gray!15}
\multicolumn{11}{c}{\textit{Turn-based AR Models}} \\

BR-Voice
& 4.87 & 4.71 & 4.66
& 74.3 & 85.7 & 96.0 & 90.4
& 83.2 & 99.8
& \textbf{90.5} \\

LFG-3
& 4.73 & 4.40 & 4.45
& 78.1 & 85.5 & 94.7 & 92.2
& 88.5 & 98.3
& 89.9 \\

Nemotron-3 Omni
& 4.75 & 4.57 & 4.58
& 71.4 & 82.3 & 93.0 & 91.1
& 88.7 & 100
& 89.4 \\

Qwen3-Omni-Thinking
& 4.82 & 4.53 & 4.53
& 78.1 & 83.0 & 94.3 & 88.9
& 80.6 & 97.2
& 88.8 \\

GPT-4o-Audio
& 4.78 & 4.49 & 4.58
& 75.5 & 80.3 & 89.2 & 84.1
& 76.0 & 98.7
& 86.8 \\

Kimi-Audio
& 4.46 & 3.97 & 4.20
& 63.1 & 62.2 & 83.5 & 69.7
& 61.1 & 100
& 76.9 \\

\rowcolor{gray!15}
\multicolumn{11}{c}{\textit{Turn-based Diffusion Models}} \\

DIFFA-2
& 3.86 & 3.67 & 2.80$^*$
& 40.8 & 38.1 & 61.5 & 53.1$^*$
& 40.8 & 85.6
& 58.5 \\

\rowcolor{gray!15}
\multicolumn{11}{c}{\textit{Full-duplex AR Models}} \\

Nemotron-3 VoiceChat
& 3.59 & 3.12 & 3.06
& 43.4 & 50.5 & 64.4 & 52.7
& 17.1 & 99.6
& 58.1 \\

Lychee-FD
& 3.31 & 2.28 & 1.75 
& 34.4 & 22.9 &29.2 & 51.0
& 22.6 & 84.4 & 
43.5

\\

Freeze-Omni
& 4.03 & 3.46 & 3.15
& 53.5 & 28.1 & 31.0 & 50.7
& 23.4 & 97.3
& 55.2 \\

Moshi
& 2.01 & 1.60 & 1.30
& 15.6 & 24.0 & 25.9 & 47.4
& 10.1 & 44.2
& 29.5 \\

\rowcolor{gray!15}
\multicolumn{11}{c}{\textit{Stage-2}} \\

Diffuplex \textit{w/o} thinking
& 4.71 & 4.36 & 4.48 & 60.2 & 77.4 & 89.7 & 83.4 & 77.2 & 97.5 & 84.1 \\

Diffuplex \textit{w.} thinking 
& 4.70 & 4.32 & 4.47 & 63.8 & 83.0 & 94.1 & 88.4 & 84.4 & 98.7 & 86.9 \\

\rowcolor{gray!15}
\multicolumn{11}{c}{\textit{Stage-3}, $K{=}4$} \\

Diffuplex (\textit{full-duplex})
& 4.65 & 4.29 & 4.30
& 59.3 & 71.9 & 85.9 & 83.1
& 75.6 & 100.0 & 82.3 \\

\bottomrule
\end{tabular}
}
\end{table*}

As shown in Table~\ref{tab:voicebench}, the Stage~2 response generator scores 84.1 without thinking and 86.9 with thinking. The largest gains from enabling thinking occur on IFEval ($+7.2$), MMSU ($+5.6$), BBH ($+5.0$), and OpenBookQA ($+4.4$), while the three open-ended QA subsets change little. This suggests that additional inference-time reasoning helps most on instruction following and knowledge-intensive questions. In the full-duplex Stage~3 setting, the non-thinking score is 82.3, 1.8 points below Stage~2. The largest decreases are on MMSU ($-5.5$), OpenBookQA ($-3.8$), and WildVoice ($-0.18$ on the 5-point judge scale, equivalent to $-3.6$ on the 0--100 scale), while BBH decreases by 0.3 points. Because Stage~3 freezes the response generator, this decrease does not indicate forgetting from further language-model training. It is consistent with the change to streaming turn decisions and early-prefix decoding, although these contributions are not isolated by this comparison. Appendix~\ref{app:prefix-sensitivity} shows the measured latency--quality trade-off as $K$ varies with $n{=}8$. Stage~3 still exceeds the strongest full-duplex AR baseline in the table by 24.2 points.

\textbf{Generality across diffusion backbones.} We apply the same speech adaptation and text-anchored training strategy to Nemotron-Labs-Diffusion-8B~\citep{fu2026nemotronlabsdiffusion}. The resulting speech-conditioned model reaches a VoiceBench overall score of approximately 78.7 with thinking enabled. This result shows that the strategy transfers to a second diffusion backbone; we attribute the lower score primarily to its weaker text-model capability relative to DiffusionGemma.

\subsection{Results on Conversation Behavior}
\label{sec:conversation_behavior}
\begin{table*}[t]
\centering
\caption{
Conversation behavior results.``TOR-v1" and ``TOR-v3" denote the Full-Duplex-Bench v1 and v3 (with disfluency), respectively. ``Syn.'' and ``Cand.'' stand for Synthetic and Candor audio. Closed-source models (e.g., GPT-Realtime) are excluded due to API-induced latency overheads.
}
\label{tab:fullduplexbench}
\setlength{\tabcolsep}{6pt}
\resizebox{\textwidth}{!}{
\begin{tabular}{l|ccc|cc|cc}
\toprule
\multirow{2}{*}{Model}
& \multicolumn{3}{c|}{Turn Taking}
& \multicolumn{2}{c|}{User Interruption}
& \multicolumn{2}{c}{Pause Handling} \\
& TOR-v1 $\uparrow$ & TOR-v3 $\uparrow$ & Latency $\downarrow$
& TOR $\uparrow$ & Latency $\downarrow$
& Syn. $\downarrow$ & Cand. $\downarrow$ \\
\midrule

Freeze-Omni
& 0.336 & 0.312 & 0.953
& 0.867 & 1.409
& 0.642 & 0.481 \\

Moshi
& 0.941 & 0.925 & 0.265
& 1.000 & 0.257
& 0.985 & 0.980 \\

PersonaPlex
& 0.992 & 0.985 & 0.070
& 1.000 & 0.400
& 0.584 & 0.662 \\

Raon-SpeechChat
& 0.832 & 0.810 & 1.034
& 0.980 & 1.219
& 0.212 & 0.299 \\

Nemotron-VC
& 0.820 & 0.798 & 0.448
& 1.000 & 0.480
& 0.153 & 0.255 \\

Lychee-FD
& 0.983 & 0.971 & 0.840
& 0.945 & -
& 0.100 & - \\ \midrule

Diffuplex 
& 0.933 & 0.980 & 0.396  &  0.945
& 0.239 & 0.679  
&  0.667 \\
Diffuplex (\textit{w/o} Fisher)
& 1.000 & 0.990 & 0.536 & 1.000
& 0.691  & 0.380
& 0.431  \\

\bottomrule
\end{tabular}
}
\end{table*}
We report two variants of our model because we find that adding the Fisher conversational corpus~\citep{cieri2004fisher} into Stage 3 training noticeably changes the learned conversation policy. Including Fisher favors aggressive, low-latency turn transitions; without Fisher, the model focuses on semantically informed turn-taking. The remaining baselines are representative full-duplex spoken dialogue models proposed in recent years. We evaluate all systems on turn-taking, user interruption, and pause handling, where higher TOR and lower latency or pause-handling scores indicate better performance. First-audio latency spans the Stage~3 AR turn decision, response denoising, and TTS. \par
As shown in Table~\ref{tab:fullduplexbench}, Diffuplex achieves the strongest turn-taking accuracy on both benchmark versions while correctly responding to user interruptions. Diffuplex with Fisher substantially reduces both turn-taking and interruption latency, but this responsiveness comes at the cost of weaker pause handling. A similar pattern appears in Moshi and PersonaPlex, both of which use Fisher-style conversational data: they enter turns rapidly, yet are more likely to interpret a temporary user pause as the end of a turn. These results expose a fundamental trade-off between aggressive response timing and semantic turn understanding. Training on fast-paced conversational data encourages the model to respond to acoustic silence, whereas excluding such data allows Diffuplex to wait for stronger semantic evidence before taking the floor. Our two variants therefore provide distinct operating points between interaction latency and robustness to natural hesitation and disfluency.

\subsection{TPF and TPS Analysis}
\label{sec:tpf}
\begin{figure*}[t]
    \centering
    \includegraphics[
        width=1.0\textwidth
    ]{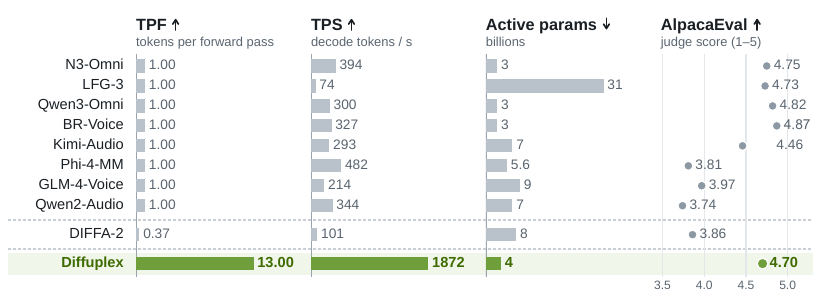}
   \caption{Decoding efficiency and score context on AlpacaEval. TPF and TPS are measured at batch~1 on one NVIDIA GB300 GPU.}
    \label{fig:efficiency}
\end{figure*}

We measure decoding efficiency on spoken AlpacaEval, whose relatively long responses make generation differences easier to observe. Figure~\ref{fig:efficiency} places tokens finalized per forward pass (\textbf{TPF}) and decoded tokens per second (\textbf{TPS}) alongside active parameter counts and AlpacaEval scores. All models use the same audio and system/user prompts, batch~1, bf16 weights and KV cache, on one NVIDIA GB300 GPU. We enable thinking where available; DIFFA-2 has no thinking mode. Appendix~\ref{app:prefix-sensitivity} explains our choice of measurement hardware.

Figure~\ref{fig:efficiency} shows that Diffuplex finalizes 13.0 tokens per forward pass, compared with 1.00 for every evaluated AR baseline and 0.37 for DIFFA-2. Its throughput reaches up to 1872~tokens/s, versus 394 for Nemotron-3 Omni and 300 for Qwen3-Omni-Thinking, while requiring 4B active parameters compared with 3B for each of those two models. The accompanying AlpacaEval scores place this efficiency comparison alongside spoken response quality. The throughput gain is smaller than the TPF gain because each diffusion pass processes a 256-token canvas and costs more than a single-token AR step. Thus, the figure shows both the pass-level parallelism and the realized decoding speed, with model size and response quality visible in the same comparison.

\subsection{Ablation and More Results}
\label{sec:ablation}
We ablate three components that are central to transferring the capabilities of a text DLM to spoken interaction. Removing Stage~1 tests whether explicit speech--text alignment is necessary before instruction tuning, while removing the text anchor evaluates whether the frozen text teacher prevents semantic drift during speech-conditioned adaptation. We additionally freeze the DLM throughout Stage~2 to examine whether adapting only the speech encoder and projection layer is sufficient. The text-input setting serves as an upper-bound diagnostic that separates the model's underlying linguistic capability from errors introduced by speech conditioning.  \par
As shown in Table~\ref{tab:ablation}, the text-input model reaches an overall score of 90.1, while Diffuplex retains 84.1 with spoken input, indicating that most of the underlying language capability is successfully transferred to speech. Removing Stage~1 causes performance to collapse to 29.9, confirming that instruction tuning alone cannot reliably ground the DLM in acoustic representations. Removing the text anchor also reduces the score substantially, from 84.1 to 66.1, demonstrating its importance in preserving the response distribution of the pretrained text model. Keeping the DLM frozen yields only 64.7, suggesting that frontend alignment alone is insufficient and that controlled adaptation of the language backbone remains necessary for strong spoken instruction following. \par

We clarify that Diffuplex can also respond user spoken request in AR mode, as cross-entropy loss is computed during Stage~1 and Stage~2. Interestingly, the AR performance is even better than diffusion decoding as shown in Table~\ref{tab:ablation}. Furthermore, Appendix~\ref{app:prefix-sensitivity} separately ablates the pass cap $K$ for committing the first $n$ tokens and reports its effect on first-audio latency and VoiceBench scores. \par
\textbf{Broader capability.} We further examine the broader capabilities of Diffuplex as a language model, with the user query given either as text or as speech (Table~\ref{tab:single_model}). These capabilities include reasoning on Big Bench Audio~\citep{artificialanalysis2024bigbenchaudio,suzgun2023bbh,srivastava2023bigbench} and GPQA-Diamond~\citep{rein2024gpqa}, knowledge on Llama-Questions~\citep{nachmani2024spectron} and MMLU-Pro~\citep{wang2024mmlupro}, and instruction following on IFEval~\citep{zhou2023ifeval}, IFBench~\citep{pyatkin2025ifbench}, and MultiChallenge~\citep{deshpande2025multichallenge}.

\begin{table*}[t]
\centering
\caption{
Ablation study on VoiceBench. ``\textit{text input}" denotes that user input is text as the test set has paired transcript. ``\textit{AR decoding}" denotes the response is generated by autoregressive decoding. Thinking is disabled for all models.
}
\label{tab:ablation}
\setlength{\tabcolsep}{4.5pt}
\resizebox{\textwidth}{!}{
\begin{tabular}{l|ccc|cccc|cc|c}
\toprule
\multirow{2}{*}{Setting}
& \multicolumn{3}{c|}{Open-ended QA}
& \multicolumn{4}{c|}{Knowledge and Reasoning}
& \multicolumn{2}{c|}{Ins. and Safety}
& \multirow{2}{*}{Overall} \\
& Alp. & Com. & Wil.
& SDQA & MMSU & OBQA & BBH
& IFE. & Adv. & \\
\midrule

Diffuplex 
& 4.71 & 4.36 & 4.48 & 60.2 & 77.4 & 89.7 & 83.4 & 77.2 & 97.5 & 84.1 \\

Diffuplex (\textit{text input})
& 4.81 & 4.53 & 4.81
& 73.4 & 84.7 & 93.8 & 94.8
& 81.3 & 100 & 90.1 \\

Diffuplex (\textit{AR decoding})
& 4.77 & 4.38 & 4.50
& 64.6 & 73.8 & 87.9 & 86.5
& 80.3 & 99.6 & 85.1 \\

\midrule

\multicolumn{1}{r|}{\textit{w/o} text anchor}
& 3.96 & 3.25 & 3.05
& 48.4 & 54.2 & 64.6 & 61.5
& 61.3 & 100 & 66.1 \\

\multicolumn{1}{r|}{\textit{w/o} Stage~1}
& 1.00 & 1.00 & 1.01
& 0.0 & 24.3 & 31.2 & 44.8
& 15.1 & 93.8 & 29.9 \\

\multicolumn{1}{r|}{\textit{w/o} trainable DLM}
& 4.02 & 3.16 & 3.51
& 50.0 & 52.8 & 62.5 & 56.2
& 49.8 & 96.9 & 64.7 \\ 
\bottomrule
\end{tabular}
}
\end{table*}
\begin{table}[t]
\centering
\caption{Diffuplex across more benchmarks, grouped by capability.
Shaded columns are spoken (audio-input) benchmarks; the others are text.}
\label{tab:single_model}
\small
\setlength{\aboverulesep}{0pt}\setlength{\belowrulesep}{0pt}
\newcolumntype{C}{>{\centering\arraybackslash}c}
\newcolumntype{A}{>{\columncolor{audiobg}\centering\arraybackslash}c}
\renewcommand{\arraystretch}{1.35}
\newcommand{\singlemodeltab}{%
\begin{tabular}{ACACCCC}
\toprule
\multicolumn{2}{c}{\textbf{Reasoning}} &
\multicolumn{2}{c}{\textbf{Knowledge}} &
\multicolumn{3}{c}{\textbf{Instruction following}} \\
\cmidrule(l{0.6em}r{0.6em}){1-2}\cmidrule(l{0.6em}r{0.6em}){3-4}\cmidrule(l{0.6em}r{0.6em}){5-7}
\makecell{Big Bench Audio} &
\makecell{GPQA-Diamond} &
\makecell{Llama-Questions} &
\makecell{MMLU-Pro} &
\makecell{IFEval} &
\makecell{IFBench} &
\makecell{MultiChallenge} \\
\midrule
97.5 & 67.4 & 81.3 & 80.4 & 94.5 & 61.6 & 53.5 \\
\bottomrule
\end{tabular}}%
\newlength{\smtnat}%
\settowidth{\smtnat}{\setlength{\tabcolsep}{0pt}\singlemodeltab}%
\setlength{\tabcolsep}{\dimexpr(\textwidth-\smtnat)/14\relax}%
\singlemodeltab
\end{table}
 
\section{Conclusion}

We introduced Diffuplex, a full-duplex spoken language model that decouples conversational control from semantic generation. Diffuplex utilize AR decoding to continuously determine \emph{when to speak}, while a speech-conditioned DLM independently generates \emph{what to say} through parallel refinement. This separation preserves the linguistic capabilities of a strong text model without binding semantic computation to the physical audio frame rate. Together with anchored-prefix decoding and modular speech synthesis, Diffuplex enables low-latency interaction while retaining strong spoken intelligence. Experiments across spoken understanding and full-duplex dialogue benchmarks demonstrate that Diffuplex substantially narrows the capability gap between responsive full-duplex systems and intelligent turn-based models. We hope this two-axis formulation provides a useful foundation for future spoken agents that are both conversationally natural and cognitively capable.

\clearpage

\bibliographystyle{iclr2027_conference}
\bibliography{iclr2027_conference}

\appendix
\clearpage
\section{Figure~1 Details}
\label{app:fig1-details}
\begin{figure*}[t]
    \centering
    \includegraphics[width=0.95\textwidth]{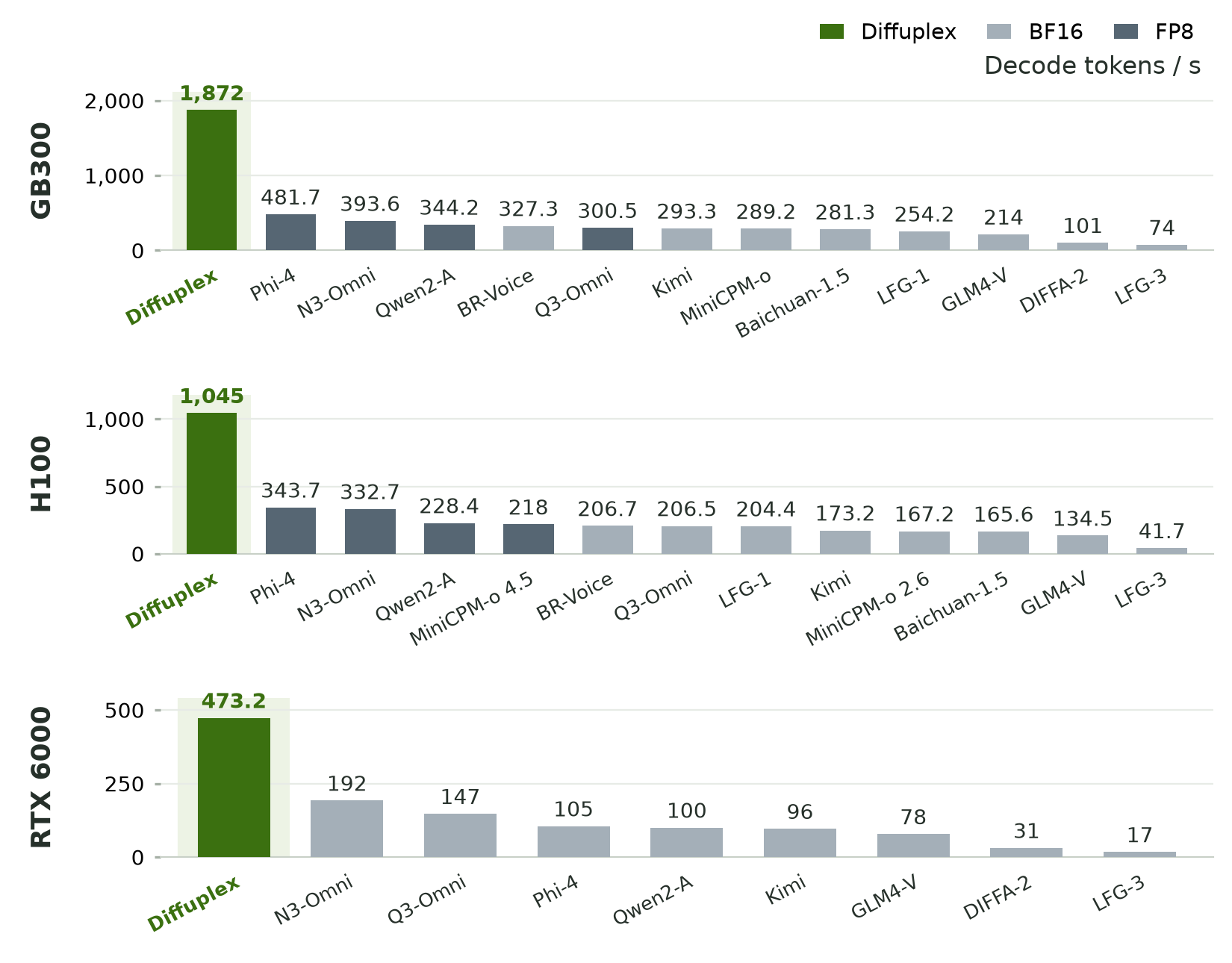}
    \caption{Output speed across different hardware (GPU). For RTX 6000, the Diffuplex is BF16 while GB300 and H100 are FP8.}
    \label{fig:3gpu}
\end{figure*}

Beyond GB300, we report the decoding speed on NVIDIA RTX 6000 and H 100 in Figure~\ref{fig:3gpu}, which are popular GPU types on cluster even local machine. The test set is AlpacaEval and the TPF ranges from 13.0 to 13.7.  

\textbf{Inference Implementation on GB300.}
\label{app:engine}
We decode with a torch-free CUDA runtime that implements the sampler described in Sec.~\ref{infer_and_denoise}. In precision, the mixture-of-experts layers use a grouped matrix multiplication with 4-bit weights and activations (NVFP4, E2M1, 16-element groups) in place of 4-bit weights with 16-bit activations, the dense projections and the
head use FP8, the rotary tables are kept in FP32, and amax reduction, scale finalization and row quantization are fused into one pass; the experts are requantized from bf16 with 9.5\% relative RMS error, and the model occupies 16.2\,GB, 3.2$\times$ less than bf16. In attention, we use the FlashAttention-4 kernels for the 256-wide head dimension. In the mixture-of-experts and dense layers, every matrix multiplication uses a per-shape tuned plan with a fixed split-$K$ quantum so that it is graph-capturable, and the expert activation scales are finalized in a fused kernel.
\section{More Discussion}
\label{app:more-discussion}

\paragraph{Can Stage~3 be trained without the listening-mode LoRA?} \par
The interaction head can be trained without the adapter, but the resulting policy mainly learns an acoustic voice-activity detector: it reacts to speech and silence while failing to distinguish a within-turn hesitation from a true turn ending. In particular, it does not handle pauses reliably. The listening-mode LoRA gives the causal speech pathway enough flexibility to use conversational context for turn decisions while the diffusion response pathway remains fixed.

\paragraph{Why is there no direct multi-token-prediction (MTP) baseline?} \par
Our efficiency metric counts finalized \emph{answer-text} tokens per backbone forward pass. Diffuplex reaches 10.15 on spoken AlpacaEval, above the fixed proposal horizons of representative text MTP configurations, such as four-token prediction~\citep{gloeckle2024mtp}; this is a comparison with those configurations, not a universal upper bound on MTP. Among the speech-input MTP studies most relevant to our setting, ParaASR applies a six-token proposal to acoustically constrained transcription~\citep{lin2026paraasr}, whereas VocalNet and PRIME-Speech apply MTP to \emph{speech or codec tokens} rather than parallelizing the answer-text stream~\citep{wang2025vocalnet,hu2026primespeech}. Their output units and tasks differ from open-ended spoken question answering, so their reported token rates cannot be placed directly alongside our answer-text TPF. A promising solution is using MTP to replace the early prefix in Diffuplex, which might result in lower first-audio latency. We view it as our future work.

\section{Training Details}
\label{app:training-details}

We train the speech-conditioned response model in three phases: Stage~1a, Stage~1b, and Stage~2. The backbone is DiffusionGemma-26B-A4B-it~\citep{diffusiongemma2026}, paired with a Parakeet speech encoder~\citep{nvidia2025parakeet} and a projection layer that maps speech features into the backbone's embedding space. These response-model phases use a 256-token diffusion canvas and an auxiliary autoregressive (AR) next-token objective. Audio is converted to 16-kHz mono waveforms before feature extraction. The training seed is 42.

\subsection{Stage 1a: Projector Warm-up}

We train the projection layer while freezing the speech encoder and language model. The data comprise the LibriSpeech~\citep{panayotov2015librispeech} train-clean-100, train-clean-360, and train-other-500 partitions, totaling 281,099 utterances (approximately 960 hours). Targets are lowercase transcripts without punctuation. We train for two epochs, or 1,098 optimizer updates, with global batch size 512. The AR loss has weight 0.1 relative to the diffusion loss. We use AdamW with peak learning rate $3\times10^{-4}$, 28 warm-up updates, and cosine decay to $3\times10^{-5}$.

\subsection{Stage 1b: Speech Encoder Adaptation}

Starting from Stage~1a, we unfreeze the speech encoder and continue training it with the projector while keeping the language model frozen. The data are a 32-source English ASR mixture drawn primarily from Granary~\citep{raokoluguri2025granary}, with additional internal ASR and LibriSpeech data. Unlike Stage~1a, the transcription targets retain punctuation and capitalization. We train for 4,032 updates with global batch size 512, corresponding to approximately 2.06 million sampled presentations. The AR loss weight is 0.05. We use AdamW with peak learning rate $2\times10^{-5}$, 403 warm-up updates, and cosine decay to $10^{-5}$.

\subsection{Stage 2: Spoken Instruction Tuning}

Starting from the Stage~1b speech frontend, Stage~2 adapts the model to spoken requests. Its training pool contains 600,000 speech-instruction examples selected from Nemotron-Omni SFT~\citep{nvidia2026nemotron3omni} and 37,500 ASR replay examples. The two tasks are sampled with probabilities 95\% and 5\%. We optimize the speech encoder, projector, and diffusion language model, keeping the MoE router frozen. The diffusion and AR loss weights are both 1.0. To retain the pretrained model's textual capabilities, a frozen copy of the language model receives the written form of each speech-instruction example; forward KL losses between teacher and student are applied to the diffusion canvas and AR readout with weights 3.0 and 1.0, respectively.

Stage~2 uses a maximum padded sequence length of 8,192 tokens and global batch size 128. We use AdamW with peak learning rates $10^{-5}$ for the speech frontend and $3\times10^{-6}$ for the language model, 200 warm-up updates, and cosine decay. The selected joint-training checkpoint is after 2,266 optimizer updates. Stages~1a, 1b, and 2 use 64 GPUs with BF16 computation, FP32 optimizer parameters, and gradient-norm clipping at 1.0.

\subsection{Stage 3: Full-duplex Interaction Training}

\paragraph{Data.}
The main Stage~3 mixture combines 120 hours of clean synthetic conversations with 70 hours of internal conversational corpus. The clean data provide clearly defined turns and pauses; the internal data add natural hesitations, backchannels, and user interruptions. We also train a variant with Fisher telephone conversations~\citep{cieri2004fisher} to expose the policy to faster, spontaneous turn exchanges. For the noise-augmented recipe, we add stationary fan and office ambience selected from DNS5~\citep{dubey2023dns} and DEMAND~\citep{thiemann2013demand} to the clean synthetic recordings. We also mix quiet pink/white noise and short nonspeech events such as doors, typing, and chair movement. A small fraction of the recordings retains its original silence.

\paragraph{Optimization.}
Starting from the Stage~2 checkpoint, we train the four-state interaction head and a listening-mode LoRA adapter~\citep{hu2021lora}, while keeping the response generator fixed. The policy is trained with cross-entropy on 80-ms state labels, with extra weight on the sparse start- and stop-speaking transitions. We use AdamW with peak learning rate $10^{-3}$, 50 warm-up updates, and cosine decay to $10^{-4}$ over 3,000 updates. Global batch size is 32 on two nodes with eight GPUs each.

\section{Compared Models and Their Sources}
\label{app:model-sources}

We cite the original reports or public model cards for every model used in the comparisons. The systems address different operating regimes, and their reported scores and latencies should be read together with the evaluation settings in Sections~\ref{sec:spoken_intelligence}--\ref{sec:tpf}.

\paragraph{Turn-based speech and audio models.}
LFG-3 is documented in its public model card~\citep{glenn2026lfg3}. Nemotron-3 Omni~\citep{nvidia2026nemotron3omni}, Qwen3-Omni-Thinking~\citep{xu2025qwen3omni}, and Kimi-Audio~\citep{kimiteam2025kimiaudio} provide the other high-capability comparisons in Table~\ref{tab:voicebench}. The Whisper-L + GPT-4o cascade uses Whisper~\citep{radford2022whisper} and GPT-4o~\citep{openai2024gpt4o}; the GPT-4o-Audio row is likewise attributed to GPT-4o's audio-capable system~\citep{openai2024gpt4o}. The efficiency comparison in Figure~\ref{fig:efficiency} additionally includes Phi-4-Multimodal~\citep{microsoft2025phi4multimodal}, GLM-4-Voice~\citep{zeng2024glm4voice}, and Qwen2-Audio~\citep{chu2024qwen2audio}. These models differ in parameter activation and speech-output design, which is why Figure~\ref{fig:efficiency} displays active parameters and a quality measure alongside decoding speed.

\paragraph{Diffusion-based models.}
DIFFA-2~\citep{zhou2026diffa2} is the speech-conditioned diffusion baseline. Our response backbone is DiffusionGemma~\citep{diffusiongemma2026}; the backbone-transfer experiment uses Nemotron-Labs-Diffusion~\citep{fu2026nemotronlabsdiffusion}. These citations credit the respective model authors independently of our speech adaptation and evaluation.

\paragraph{Full-duplex systems.}
The conversational baselines include Moshi~\citep{defossez2024moshi}, Freeze-Omni~\citep{wang2024freezeomni}, PersonaPlex~\citep{roy2026personaplex}, Lychee-FD~\citep{liu2026lychee}, Raon-SpeechChat~\citep{kim2026raonspeech}, and Nemotron VoiceChat~\citep{balam2026voicechat}; ``Nemotron-VC'' is the abbreviated label in Figure~\ref{fig:two_axes}. Figure~\ref{fig:two_axes} also includes SALM-Duplex~\citep{hu2025salmduplex} and FLAIR~\citep{wu2026flair}. Our implementation uses a Parakeet speech encoder~\citep{nvidia2025parakeet} and CosyVoice~3 speech synthesis~\citep{du2025cosyvoice3}. Each system is included to show a distinct point in the space of spoken intelligence, decoding speed, or conversational responsiveness.

\section{Sensitivity of Early Prefix Commitment}
\label{app:prefix-sensitivity}
The analysis in this chapter is based on the NVIDIA RTX~6000 rather than H100- or GB300-class server accelerators. The motivation is deployment relevance: spoken assistants may ultimately need to run under the tighter compute and power budgets of portable devices, so a workstation-class GPU offers a more accessible reference point than a large server accelerator. Notably, the deployed checkpoint is Stage 3 with \textbf{BF16} precision.\par
We fix the committed prefix length at $n{=}8$ tokens and vary the maximum number of denoising passes $K$ before that prefix is finalized. The reported Stage~3 setting uses $K{=}4$; the uncapped decoder uses the Stage~2 decoding configuration in a separate evaluation run. To isolate the effect of the decoding schedule, we supply the turn boundary and measure the time from that boundary to first audio. These measured post-trigger times include denoising and TTS but \textbf{exclude} the Stage~3 AR turn-decision delay.

\begin{figure*}[t]
    \centering
    \includegraphics[width=0.98\textwidth]{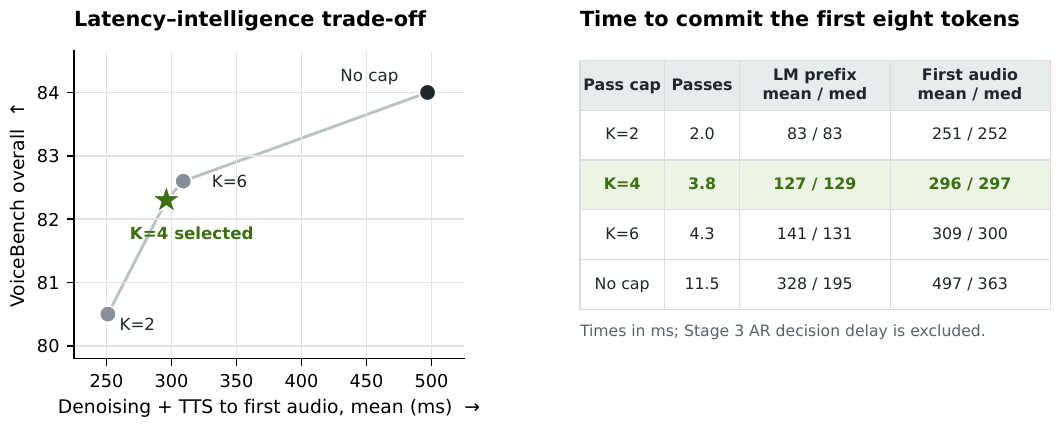}
    \caption{Measured latency--quality trade-off for committing the first $n{=}8$ tokens. Latency is measured from a supplied turn boundary to first audio, \textbf{excluding} the Stage~3 AR turn-decision delay. The left panel shows mean post-trigger latency and VoiceBench overall score; the right panel reports passes to prefix commitment and mean/median timing. The green star marks the reported $K{=}4$ configuration; ``No cap'' uses the stock decoder.}
    \label{fig:prefix-latency}
\end{figure*}

\begin{figure*}[t]
    \centering
    \includegraphics[width=0.98\textwidth]{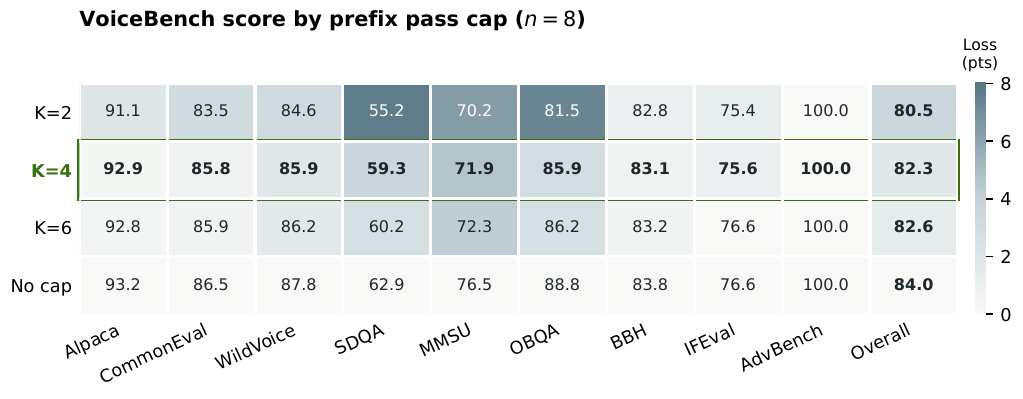}
    \caption{VoiceBench subset scores under the same $n{=}8$ prefix setting. Printed values are absolute scores on the 0--100 scale; cell shading indicates the loss relative to unconstrained decoding in the same column. The green outline identifies the $K{=}4$ setting used for the Stage~3 result.}
    \label{fig:prefix-quality}
\end{figure*}

With $K{=}2$, mean post-trigger first-audio latency is 251\,ms and the overall score is 80.5. At $K{=}4$, the first eight tokens are fixed after 3.8 passes on average; latency is 296\,ms (297\,ms median) and the score rises to 82.3. Increasing the cap to $K{=}6$ yields 309\,ms and 82.6, while uncapped decoding takes 497\,ms and scores 84.0. Across these settings, higher $K$ is associated with better VoiceBench scores and longer post-trigger latency. Figure~\ref{fig:prefix-quality} gives the corresponding subset scores.

\end{document}